# A SYSTEMATIC APPROACH FOR KINEMATIC DESIGN OF UPPER LIMB REHABILITATION EXOSKELETONS

**Rana Soltani-Zarrin, Amin Zeiaee, Reza Langari, Richard Malak**
Department of Mechanical Engineering, Texas A&M University
College Station, Texas, USA

## ABSTRACT

*Kinematic structure of an exoskeleton is the most fundamental block of its design and is determinant of many functional capabilities of it. Although numerous upper limb rehabilitation devices have been designed in the recent years, there is not a framework that can systematically guide the kinematic design procedure. Additionally, diversity of currently available devices and the many minute details incorporated to address certain design requirements hinders pinpointing the core kinematics of the available devices to compare them against each other. This makes the review of literature for identifying drawbacks of the state of the art systems a challenging and puzzling task. In fact, lack of a unifying framework makes designing rehabilitation devices an intuitive process and prone to biases from currently available designs. This research work proposes a systematic approach for kinematic design of upper limb rehabilitation exoskeletons based on conceptual design techniques. Having defined a solution neutral problem statement based on the characteristics of an ideal device, the main functionality of the system is divided into smaller functional units via the Functional Decomposition Method. Various directions for concept generation are explored and finally, it has been shown that a vast majority of the current exoskeleton designs fit within the proposed design framework and the defined functionalities.*

## INTRODUCTION

Currently, Physical Therapy (PT) and Occupational Therapy (OT) are the main two treatment options for rehabilitation of patients with movement disabilities [1]. On the other hand, robotic based therapy has proven to be an effective solution for the ever-increasing demands of rehabilitation since it can address several of the challenges facing effective rehabilitation of stroke patients [2, 3]. Robotic systems can make high intensity and customized exercises available to a large group of people, presumably at a lower cost [4]. Providing therapy to stroke patients is a labor-intensive task and this fact limits the amount of therapy that the patients can receive due to the fatigue of the therapists and the limited amount of time a therapist can spend with a patient. Considering that intensity and duration of the therapy is a determining factor for the success of rehabilitation, robotic systems can provide intensive high quality training experience to the patients while reducing the physical burden of the therapy on the therapists. One therapist can supervise many patients in the same amount of time and resultantly patients can have longer therapy sessions. Additionally, rehabilitation robots have been very promising in adding a new dimension to the type of therapies stroke patients can get. Design and utilization of games for providing therapeutic exercises and augmentation of virtual experiences such as virtual reality [5, 6] into rehabilitation devices are examples of new technologies which are aimed at improving the mental engagement of patients in therapy exercises.

Due to the many benefits rehabilitation robots can offer, there have been a surge for using robots for rehabilitation purposes. Various robotic devices have been designed for rehabilitation of upper limb in the past 20 years [7]. Design of an effective rehabilitation device requires knowledge of human anatomy, rehabilitation techniques and without any doubt, profound robotics knowledge. Kinematics of the rehabilitation robots is one of the most important and challenging aspect of their design due to the complications in modeling biological joints. As a matter of fact, designing a structure that does not limit the human range of motion while avoiding kinematic incompatibilities and the resultant hyperstaticity is a major challenge [8]. Although many upper limb rehabilitation devices have been designed in the recent years, there is not a framework that can systematically guide the kinematic design procedure. Additionally, diversity of currently available devices [7] and the many minute details incorporated to address certain design requirements hinders pinpointing the core kinematics of devices. This makes the review of literature for identifying the drawbacks of the state of the art systems a challenging and puzzling task. In fact, lack of a unifying framework makes designing rehabilitation devices an intuitive process and prone to biases from currently available designs and hinders the thorough exploration of the space of feasible design concepts.

Conceptual design techniques has long been used in the design of complicated industrial systems and commercial product development [9]. These approaches facilitate the generation of new ideas and foster inventiveness by providing a systematic methodology for design process [10]. This paper proposes a general framework for kinematic design of rehabilitation exoskeletons using conceptual design techniques. The focus of this research is on the design of exoskeletons since end effector based devices suffer from issues such as limited range of motion and uncontrolled torque transfer to the joints of patient arm and there is a consensus among rehabilitation researchers that exoskeletons surpass end effector based systems. By identifying the kinematic characteristics of an ideal exoskeleton, a solution neutral problem statement is proposed. Next, the main functionality of the system is divided into smaller functional units via the Functional Decomposition Method. Having smaller functional units that are meant to achieve a specific functionality facilitates the generation of ideas. Some guidelines are proposed for concept generation and example design concepts are generated. It has been shown that a vast majority of the current exoskeleton designs fit within the proposed design framework and the defined functionalities. In other words, we believe that solution neutral functional description of the exoskeleton's kinematics provides a means for categorizing currently available designs and identifying their drawbacks. This paper is organized as follows:

## SYSTEMATIC DESIGN METHODLOGY

Conceptual design techniques are widely used in the industry to address the multi-dimensional and multi-domain challenges of designing a new product based on the opportunities in the market, needs of the customers and the goals of the company [9]. Using a systematic design approach is advantageous due to several reasons, out of which the following are chosen due to brevity considerations. Within the systematic design frameworks, design task is seen as a process with certain steps which can effectively organize the efforts. Also, clarity of the overall process enables iterative improvement of the design within several generations of the product. Additionally, systematic design approaches enable thorough exploration of the space of feasible designs. Conceptual design techniques play an important role in achieving this by decoupling the functionalities of the system and decomposing it into smaller and more specific functionalities which can be studied independently for idea generation [11]. In fact, conceptual design is intended to provide an abstract explanation of how to achieve the desired functionality of the device. Functional modeling of the system via verb-noun pairs is the essential step for achieving such functional decomposition of the system [10].

## KINEMATIC DESIGN FRAMEWORK

As the first step of a systematic design procedure [10], the solution neutral problem statement (SNPS) is defined. By reviewing the expected functionalities and kinematic properties of an ideal rehabilitation exoskeleton in the literature, the following SNPS is proposed: "Design an exoskeleton that is kinematically compliant with human arm and does not constrain the natural motion of it". From the perspective of the kinematics, exoskeleton can be defined as an open kinematic chain that is connected to human body though more than one physical human robot interface (pHRI). It is worth mentioning that while there can be closed kinematic chains, also known as mechanisms, embedded into the structure of an exoskeleton (e.g. ARMin [12, 13], NeuroExos [14]), such devices are still an open kinematic chain as a whole. It is also important to note that the above SNPS is tailored for the kinematic design of a rehabilitation exoskeleton and thus it is only reflecting desired kinematic characteristics. Within a broader scope, other functionalities of the rehabilitation device such as its ergonomics and training capabilities might also be included in the SNPS definition.

Functional modeling of an exoskeleton is not possible by focusing on its kinematics only since the kinematic structure of an exoskeleton is a fundamental characteristic that is inseparable from the functionality of the device. Following the convention of Pahl and Beitz [9], the following functional model is proposed for the rehabilitation exoskeleton:

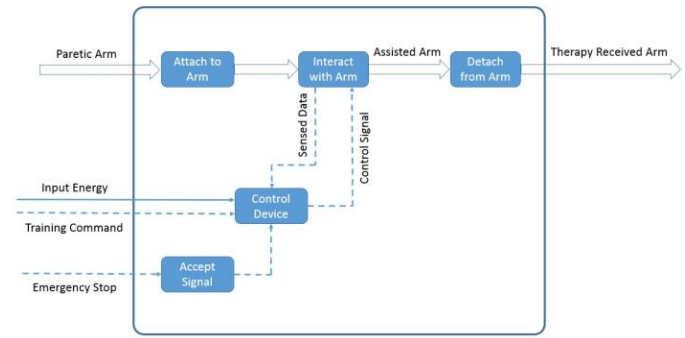

Figure 1. Functional Model of the Rehabilitation Device

The control block in the functional model of the system is responsible for actuating the device and controlling the behavior of the device respectively. "Interact with Arm" is a functional block modeling the physical interaction of the exoskeleton with human arm. Figure 2 shows the "Functional Decomposition" of this block to clarify the sub functionalities need for achieving it:

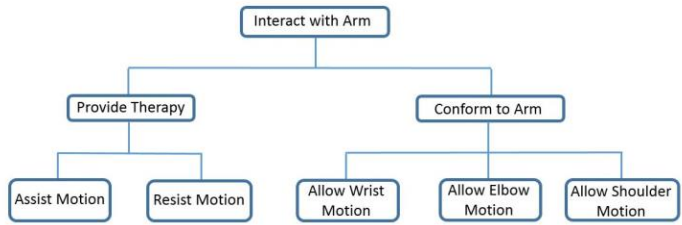

Figure 2. Functional Decomposition of the Interact with Arm Block

The interaction of the exoskeleton with the patient arm can be studied in two levels: the therapeutic and the kinematic interaction. Rehabilitation exoskeletons can be functioning in assistive (partial or full) or resistive modes depending on the higher-level training command. These sub-functions are directly related to how the device is actuated and how the force required to achieve the therapeutic goals is created. On the other hand, kinematic compatibility of the device is directly related to how the device can conform to human arm and thus can be analyzed in terms of the number of degrees of freedom in the device to support the motion of arm. In other words, "Conform to Arm" block focuses on how the device should be designed to not hinder the natural motion of the arm. The two sub functions of the "Interact with Arm" block is closely interconnected in active exoskeletons such that their classification into two sub functionalities might seem unnecessary. However, in broader scope they represent two separate functionalities. To clarify this point, consider an exoskeleton that uses functional stimulation for providing therapy to the patients and the exoskeleton frame is mainly responsible for supporting the weight of the arm and damping the motion resulted from unintended stimulation of adjacent muscle groups. In such a system, the two functionalities can be clearly distinguished.

Functional modeling is the first steps in determining how to achieve a certain functionality in an abstract level. The next step, is devising working principles for the functional blocks using the physics of the problem and major-specific knowledge.

*"Allow Shoulder Motion"*: Shoulder is one of the most complicated joints of the body to model. The motion of shoulder consists of the rotation of Humerus head in Glenoid cavity, also known as the Glenohumeral (GH) joint, as well as the motion of the so called "shoulder girdle" or "inner shoulder". The shoulder girdle is a closed kinematic chain which consists of the Scapula, Clavicle and three joints (Sternoclavicular (SC), Acromioclavicular (AC) and Scapulothoracic (ST) joints) which connect Scapula and Clavicle to each other and the rib cage. The net contribution of the shoulder girdle's complicated motion is displacement of the GH joint center in 3D space which contributes to the large range of motion of the shoulder. Thus, the functionality of "allow shoulder motion" can be achieved by generating ideas for GH joint and the inner shoulder.

Glenohumeral joint can be accurately modeled as a ball-socket joint with three degrees of freedom. Conventional spherical joint designs for robotics wrists cannot be useful since their center of rotation is within the second joint. Therefore, the objective is achieving three rotational degrees of freedom about a point that lies outside the structure. This point should be collocated with the position of human's anatomical shoulder joint. Three rotational degrees of freedom can be achieved by three consecutively connected rotary joints whose axis intercept at a single point. Figure 3 shows a schematic of such a structure:

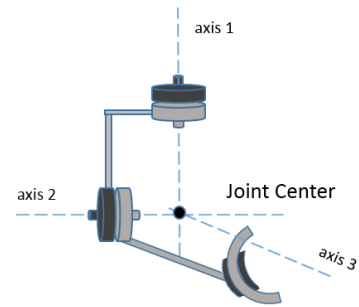

Figure 3 An Example for GH Joint Design

Various concepts with the same functionality can be generated by changing different properties of the 3 DOF joint structure in Figure 3. Examples of properties that can be modified are:

a. Use of biologic or non-biologic axes of rotations: Axes of rotations associated with the abduction/adduction (ABD), flexion/extension (FE), internal/external (IE) rotation and horizontal abduction/adduction (HABD) motions of the arm are biologic axes of the shoulder. A GH joint design might use biologic or non-biologic axes. Examples of the designs that use biologic axes are ARMin, IntelliArm exoskeletons, SUEFUL-7 [15], LIMPACT [16], Dampace [17, 18], T-Wrex exoskeletons [19]. On the hand CADEN [20], MGA [21], ARAMIS [22], CLEVERarm [23] and SAM exoskeletons [25] use non-biologic axis of rotation for shoulder.

b. The order by which 1 DoF joints are connected to each other and to the rest of device kinematic chain. For example, in Figure 3, biologic axes of rotations are used and the sequence are (HABD, FE, IE). This sequence of rotation is very popular and used in ARMin [12] and IntelliArm [26] exoskeletons. Another example of the shoulder joint axis sequence could be (HABD*, ABD*, FE*) which is used in MGA exoskeleton [21] where the asterisk is used to demonstrate the fact that these axes are not exact biologic axes and are achieved by tilting the biologic axes.

c. The angle between the consecutive axes of rotations. For example, Figure 4 shows two possibilities for the angle between the consecutive axes:

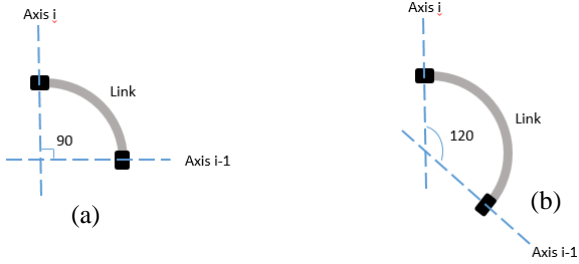

Figure 4 Angle between Consecutive Axes of Rotations as a Design Parameter

To be able to model an ideal spherical joint that creates a full sphere in 3D space the consecutive angles of rotation should satisfy [21]:

$$\begin{cases} \pi/2 - \theta_3 \leq \theta_2 \leq \pi/2 + \theta_3 \\ \pi - \theta_2 - \theta_3 \leq \theta_1 \leq \theta_2 + \theta_3 \end{cases} \quad (1)$$

where $\theta_1$, $\theta_2$ and $\theta_3$ are defined as:

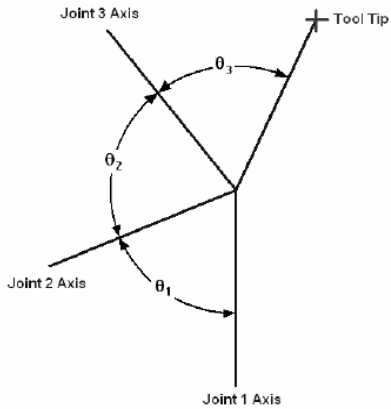

Figure 5. Angle between the Consecutive Axes of Rotations [21]

For example, $\theta_1 = \theta_2 = 90°$ and $\theta_3 = 0°$ in ARMin and $\theta_1 = 90°$, $\theta_2 = 90°$, $\theta_3 = 45°$ in MGA exoskeleton.

d. The shape of the links: The links could have any shapes as long as they preserve the requirement on the angle between the axes. Figure 3 shows a concept with piecewise linear links, while Figure 4 shows circular links.

These four methods are examples of how various GH joint designs can be achieved. Figure 6 shows some of the shoulder designs in the current exoskeletons in the literature. The design concept for GH joint in each design is enclosed in a dashed blue rectangle in each design.

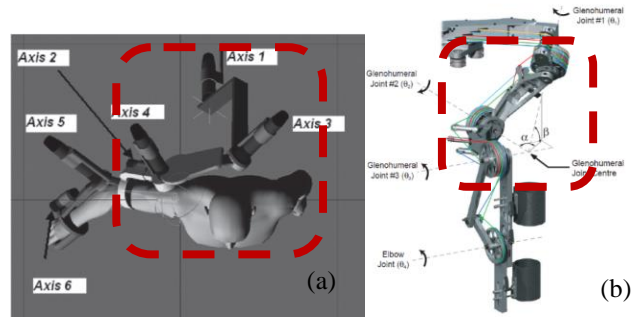

Figure 6. Shoulder Designs in: (a) ARAMIS, (b) MEDARM

As mentioned earlier, the final outcome of the shoulder girdle motion is the change of location of GH joint center. Since the center of the GH joint is the point of interception of its axes, the functionality of shoulder girdle can be achieved via two methods:

a. Manipulate axes of rotations individually making sure they are still intersecting at a single point. ARMin II exoskeleton uses this strategy for achieving the functionality of the inner shoulder. With the aid of a linkage mechanism, the elevation of arm moves the FE and IE axes of rotations vertically in ARMin II design and thus the point of interception of all three axes moves on the HABD axis vertically [12]. Figure 7 shows, the simplified shoulder mechanism of ARMin II and another example of how this method can be used.

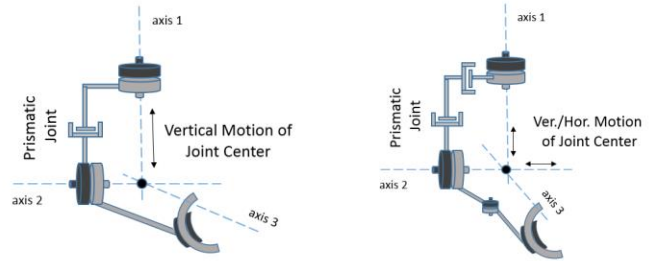

Figure 7. (a) Kinematics of ARMin II shoulder (b) Another Design Concept

b. Translating all three axes together: In this case one, two or even three degrees of freedom can be used for positioning the shoulder joint center in the 3D space. Also, various architectures of these degrees of freedom such as 3D-Cartesian structure (3 prismatic), Polar structure (1 revolute and 1 prismatic) and a single revolute joint can be used. These architectures are demonstrated in figure 8. In this figure, the GH joint concept is shown as a block which is being positioned by the proposed concepts. Using a rotary degree of freedom (Figure 8.a), the MGA exoskeleton models and follows the motion of inner shoulder on a circular path on the frontal plane of the body. Similarly, CLEVERarm has used the combination of a rotary and linear motion to support the inner shoulder motion on the frontal plane [24]. IntelliArm and Dampace exoskeleton have used the Cartesian structure (Figure 8.c).

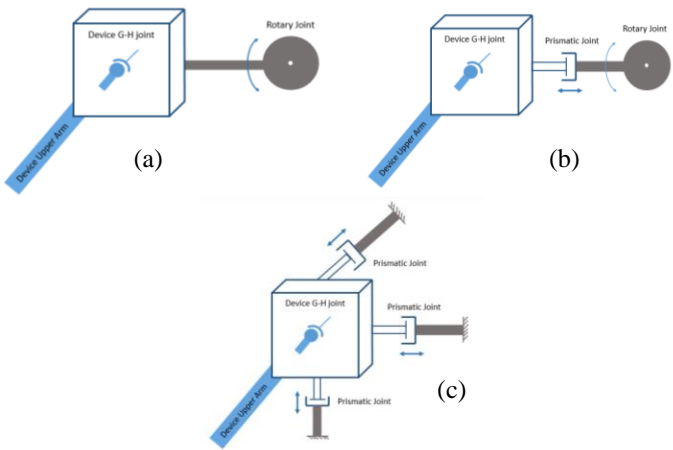

Figure 8. Design Concepts inner shoulder: (a) Single Revolute Joint (MGA), (b) Polar Structure (CLEVERarm), (c) Cartesian Structure.

"Allow Elbow Motion": Flexion/extension is the main motion of elbow joint. While a significant majority of studies model this motion with a 1 degree of freedom hinged joint, anatomical studies show that the axis of rotation of elbow joints rotates during flexion/extension which can be roughly described as a "loose" hinge joint. To be more precise, during the flexion–extension motion the elbow rotation axis traces the surface of a double quasi-conic frustum with an elliptical cross section [14] as shown in Figure 9.a. An example of a design concept for the "loose hinge" behavior of the elbow joint is shown in Figure 9.b. This concept is used in NeuroExos exoskeleton where the universal joint is coupled with the elbow flexion/extension and actualized through a complicated linkage design:

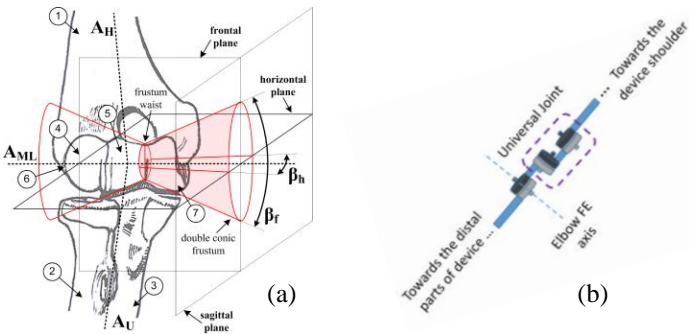

Figure 9. (a) Anatomy of Elbow and Rotation of the Elbow Axis [13], (b) A Design Concept to Allow this Motion

"Allow Hand Motion": Hand can achieve three different rotations, namely wrist flexion/extension (WFE), radial/ulnar deviation (UD) and pronation/supination (PS), of which the two formers are realized by the wrist joint. The simplest model for the two rotations of the wrist would be a universal joint; however, this might not be the most accurate model of the wrist since it has been shown that the radial/ulnar deviation axis is not exactly on the wrist and is located at the distal end of the forearm. While some studies consider pronation/supination as one of the degrees of freedom that the wrist provides, the authors believe that such categorization is inaccurate since pronation/supination motion is resulted from the motion of Radius around Ulna within the upper arm. Therefore, pronation/supination can be achieved with a rotation around the axial direction of the forearm. However, the complexity arises due to the so called "forearm load angle", which is the lateral angle between the axis of the Humerus and the forearm. Figure 10 shows example design concepts for achieving forearm PS:

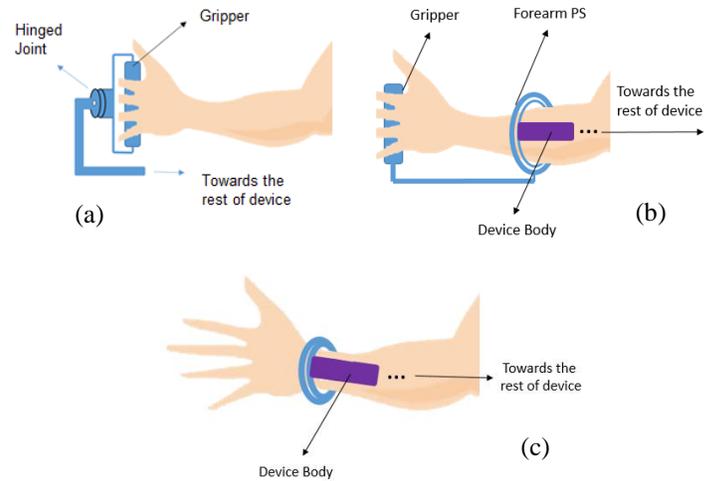

Figure 10. Example Concepts for Forearm PR: (a) MGA, (b) ARMin II, (c) Limpact

It is worth mentioning that not all the currently available exoskeletons have all the degrees of freedom discussed above. For example, T-wrex and ARMin II exoskeleton do not support the IE rotation of the shoulder and wrist WFE respectively. In fact, almost all of the functioning exoskeletons only support a subset of the main degrees of freedom in the human arm. This is mainly because addition of degrees of freedom makes the design heavier and more complex.

The other unit in the functional model of the exoskeleton in Figure 1 is the "Attach to Arm" block which describes the physical interface between the human arm and the exoskeleton. Examples of such physical interfaces are the exoskeleton gripper or the coughs that are used to attach the device to the paretic arm. Examples of the design parameters to be considered are the number of these interface points, which part of the arm they connect to the device and the rigidity of the connection between the body and the device. As mentioned earlier, exoskeleton is an open kinematic chain as a whole, within which the functional units of Figure 2 are embedded. Therefore, interconnection of these blocks within the kinematic chain is also important to be studied. Adding additional degrees of freedom within these interconnections or at the physical interface points give rise to the self-aligning exoskeleton such as Limpact, Dampace and ABLE [27] exoskeletons.

## CONCLUSION

Kinematic structure of an exoskeleton is the most fundamental block of its design and is determinant of many functional capabilities of it. Although numerous upper limb rehabilitation devices have been designed in the recent years, there is not a framework that can systematically guide the kinematic design procedure. This research work proposes a systematic approach for kinematic design of upper limb rehabilitation exoskeletons based on conceptual design techniques. Having defined a solution neutral problem statement, the main functionality of the system is divided into smaller functional units via the Functional Decomposition Method. Various directions for concept generation are explored and finally, it has been shown that a vast majority of the current exoskeleton designs fit within the proposed design framework and the defined functionalities